\documentclass[twocolumn]{webofc}
\usepackage[varg]{txfonts}   
\usepackage{algorithm}
\usepackage{algorithmic}
\usepackage{subfigure}
\usepackage{multirow}
\usepackage{threeparttable}
\usepackage{booktabs}
\begin{document}
\title{Porcellio scaber algorithm (PSA) for solving constrained optimization problems}
\author{\firstname{Yinyan} \lastname{Zhang}\inst{1}\fnsep\thanks{\email{yinyan.zhang@connect.polyu.hk}} \and
        \firstname{Shuai} \lastname{Li}\inst{1}\fnsep\thanks{\email{shuaili@polyu.edu.hk}} \and
        \firstname{Hongliang} \lastname{Guo}\inst{2}\fnsep\thanks{\email{guohl1983@uestc.edu.cn}}
}

\institute{Department of Computing, The Hong Kong Polytechnic
University, Hong Kong, China \and
           Center for Robotics, School of Automation Engineering,
University of Electronic Science and Technology of China, Chengdu,
China}

\abstract{%
  In this paper, we extend a bio-inspired algorithm called the
porcellio scaber algorithm (PSA) to solve constrained optimization
problems, including a constrained mixed discrete-continuous
nonlinear optimization problem. Our extensive experiment results
based on benchmark optimization problems show that the PSA has a
better performance than many existing methods or algorithms. The
results indicate that the PSA is a promising algorithm for
constrained optimization. }
\maketitle

\section{Introduction}
Modern optimization algorithms may be roughly classified into
deterministic optimization algorithms and stochastic ones. The
former is theoretically sound for well-posed problems but not
efficient for complicated problems. For example, when it comes to
nonconvex or large-scale optimization problems, deterministic
algorithms may not be a good tool to obtain a globally optimal
solution within a reasonable time due to the high complexity of the
problem. Meanwhile, while stochastic ones may not have a strong
theoretical basis, they are efficient in engineering applications
and  have become popular in recent years due to their capability of
efficiently solving complex optimization problems, including NP-hard
problems such as the travelling salesman problem. Bio-inspired
algorithms take an important role  in stochastic algorithms for
optimization. These algorithms are designed based on the
observations of animal behaviors. For example, one of the well known
bio-inspired algorithm called particle swarm optimization initially
proposed by Kennedy and Eberhart \cite{1} is inspired by the social
foraging behavior of some animals such as the flocking behavior of
birds.

There are some widely used benchmark problems in the field of
stochastic optimization. The pressure vessel design optimization
problem is an important benchmark problem in structural engineering
optimization \cite{2}. The problem is a constrained mixed
discrete-continuous nonlinear optimization problem. In recent years,
many bio-inspired algorithms have been proposed to solve the problem
\cite{aipso,csaam,doose,eoclb}.  The widely used benchmark problems
also include a nonlinear optimization problem proposed by Himmelblau
\cite{anlp}.

Recently, a novel bio-inspired algorithm called the porcellio scaber
algorithm (PSA) has been proposed by Zhang and Li \cite{PSA}, which
is inspired by two behaviors of porcellio scaber. In this paper, we
extend the result in \cite{PSA} to solve constrained optimization
problems. As the original algorithm proposed in \cite{PSA} deals
with the case without constraints, we provide some improvements for
the original PSA so as to make it capable of solving constrained
optimization problems. Then, we compare the corresponding experiment
results with reported ones for the aforementioned benchmark problems
as case studies. Our extensive experiment results show that the PSA
has much better performance in solving optimization problems than
many existing algorithms. Before ending this introductory section,
the main contributions of this paper are listed as follows:
\begin{itemize}
\item[1)] We extend the PSA to solve constrained optimization problems, including the constrained mixed
discrete-continuous nonlinear optimization problem.
\item[2)] We show that the PSA is better than many other existing algorithms in
solving constrained optimization problems by extensive numerical
experiments.
\end{itemize}

\begin{algorithm}[t]
\caption{Original PSA}
\begin{algorithmic}
\STATE {\it Cost function} $f(\mathbf{x})$,
~$\mathbf{x}=[x_1,x_2,\cdots,x_d]^\text{T}$ \STATE {\it Generate
initial position of porcellio scaber}
$\mathbf{x}^0_i~(i=1,2,\cdots,N)$ \STATE {\it Environment condition}
$E_{\mathbf{x}}$ {\it at position} $\mathbf{x}$ {\it is determined
by} $f(\mathbf{x})$ \STATE {\it Set weighted parameter} $\lambda$
{\it for decision based on aggregation and the propensity to explore
novel environments}\STATE {\it Initialize $f_{*}$ to an extremely
large value} \STATE {\it Initialize each element of vector
$\mathbf{x}_{*}\in\mathbb{R}^d$ to an arbitrary value}
\WHILE{$k<MaxStep$} \STATE {\it Get the position with the best
environment condition, i.e.,}
$\mathbf{x}_b=\text{arg}\min_{\mathbf{x}^k_j}\{f(\mathbf{x}^k_j)\}$
{\it at the current time among the group of porcellio scaber}
\IF{$\min_{\mathbf{x}^k_j}\{f(\mathbf{x}^k_j)\}<f_*$} \STATE
$\mathbf{x}_*=\mathbf{x}_b$ \STATE
$f_*=\min_{\mathbf{x}^k_j}\{f(\mathbf{x}^k_j)\}$\ENDIF \STATE {\it
Randomly chose a direction
$\tau=[\tau_1,\tau_2,\cdots,\tau_d]^\text{T}$ to detect} \STATE {\it
Detect the best environment condition} $\min\{E_{\mathbf{x}}\}$ {\it
and worst environment condition} $\max\{E_{\mathbf{x}}\}$ {\it at
position} $\mathbf{x}^{k}_i+\tau$ {\it for} $i=1:N$ {\it all} $N$
{\it porcellio scaber} \FOR{$i=1:N$~{\it all $N$ porcellio scaber}}
\STATE {\it Determine the difference with respect to the position to
aggregate i.e.,}
$\mathbf{x}^k_i-\text{arg}\min_{\mathbf{x}^k_j}\{f(\mathbf{x}^k_j)\})$
 \STATE {\it Determine where to explore, i.e.}, $p\tau$
 \STATE {\it Move to a new position according to (\ref{eq.formula})} \ENDFOR \ENDWHILE \STATE{\it Output $\mathbf{x}_*$ and
the corresponding function value $f_*$} \STATE {\it Visualization}
\end{algorithmic}
\end{algorithm}

\section{Problem Formulation}\label{sec.2}
The constrained optimization problem (COP) considered in this paper
is presented as follows:
\begin{equation}\label{pro}
\begin{aligned}
\text{minimize}~\check{f}(\mathbf{x}),& \\
\text{subject to}~g_j(\mathbf{x})&\leq0,\\
l_i\leq x_i&\leq u_i,\\
\end{aligned}
\end{equation}
with $i=1,2,\cdots,d$ and $j=1,2,\cdots,m$, where
$\mathbf{x}=[x_1,x_2,\cdots,x_d]^\text{T}$ is a $d$-dimension
decision vector; $l_i$ and $u_i$ are the corresponding lower bound
and upper bound of the $i$th decision variable;
$\check{f}(\mathbf{x}):\mathbb{R}^d\rightarrow \mathbb{R}$ is the
cost function to be minimized. For the case that the problem is
convex, there are many standard algorithms to solve the problem.
However, for the case that the problem is not convex, the problem is
difficult to solve.

\section{Algorithm Design}\label{sec.3}
In this section, we modify the original PSA \cite{PSA} and provide
an improved PSA for solving COPs.

\subsection{Original PSA}
For the sake of understanding, the original PSA is given in
algorithm 1 \cite{PSA}, which aims at solving unconstrained
optimization problems of the following form:
$$\text{minimize}~{f}(\mathbf{x}),$$ where $\mathbf{x}$ is the
decision vector and $f$ is the cost function to be minimized.
 The main formula of the original PSA is
given as follows \cite{PSA}:
\begin{equation}\label{eq.formula}
{\mathbf{x}^{k+1}_i=\mathbf{x}^{k}_i-(1-\lambda)(\mathbf{x}^k_i-\text{arg}\min_{\mathbf{x}^k_j}\{f(\mathbf{x}^k_j)\})+\lambda
p\tau},
\end{equation}
where $\lambda\in(0,1)$, $\tau$ is a vector with each element being
a random number, and $p$ is defined as follows:
\begin{equation*}
p=\frac{f(\mathbf{x}^k_i+\tau)-\min\{f(\mathbf{x}^k_i+\tau)\}}{\max\{f(\mathbf{x}^k_i+\tau)\}-\min\{f(\mathbf{x}^k_i+\tau)\}}.
\end{equation*}
Evidently, the original PSA does not take constraints into
consideration. Thus, it cannot be directly used to solve COPs.

\subsection{Inequality constraint conversion}
In this subsection, we provide some improvements for the original
PSA and make it capable of solving COPs. As the original PSA focuses
on solving unconstrained problem, we first incorporate the
inequality constraints $g_j(\mathbf{x})\leq0$ ($j=1,2,\cdots,m$)
into the cost function. To this end, the penalty method is used, and
a new cost function is obtained as follows:
\begin{equation}\label{eq.checkf}
\begin{aligned}
\check{f}(\mathbf{x})&=f(\mathbf{x})+\gamma\sum_{i=1}^{m}g^2_i(\mathbf{x})h(g_i(\mathbf{x})),
\end{aligned}
\end{equation}
where $h(g_i(\mathbf{x}))$ is defined as
\begin{equation*}
h(g_i(\mathbf{x}))=\begin{cases} 1,~\text{if}~g_i(\mathbf{x})>0,\\
0,~\text{if}~g_i(\mathbf{x})\leq0,
\end{cases}
\end{equation*}
and $\gamma\gg1$ is the penalty parameter. By using a large enough
value of $\gamma$ (e.g., $10^{12}$), unless all the inequality
constraints $g_i(\mathbf{x})\leq0$ ($i=1,2,\cdots,m$) are satisfied,
the term $\gamma\sum_{i=1}^{N}g^2_i(\mathbf{x})h(g_i(\mathbf{x}))$
takes a dominant role in the cost function. On the other hand, when
all the inequality constraints $g_i(\mathbf{x})\leq0$
($i=1,2,\cdots,m$) are satisfied, $h(g_i(\mathbf{x}))=0$, $\forall
i$, and thus $\check{f}(\mathbf{x})=f(\mathbf{x})$.

\subsection{Addressing simple bounds}
In terms of the simple bounds $l_j\leq x_j\leq u_j$ with
$j=1,2,\cdots,d$, they are handled via two methods. Firstly, to
satisfy the simple bounds,  the initial position of each porcellio
scaber is set via the following formula:
\begin{equation}\label{eq.ini}
\begin{aligned}
{x}^0_{i,j}=l_j+(u_j-l_j)\times rand(0,1)
\end{aligned}
\end{equation}
where ${x}^0_{i,j}$ denotes the initial value of the $j$th  variable
of the position vector of the $i$th (with $i=1,2,\cdots,N$)
porcellio scaber; ${rand}(0,1)$ denotes a random number in the
region $(0,1)$, which can be realized by using the  ${rand}$
function in Matlab. The formula (\ref{eq.ini}) guarantees that the
initial positions of all the porcellio scaber satisfy the the simple
bounds $l_j\leq x_j\leq u_j$ with $j=1,2,\cdots,d$.

Secondly, if the positions of all the  porcellio scaber are updated
according to \eqref{eq.formula} by replacing $f(\mathbf{x})$ with
$\check{f}(\mathbf{x})$ defined in (\ref{eq.checkf}) for the
constrained optimization problem (\ref{pro}), then the updated
values of the position vector $\mathbf{x}^k_i$ may violate the
simple bound constraints. To handle this issue, based on
\eqref{eq.formula}, a modified evolution rule is proposed as
follows:
\begin{equation}\label{eq.mod}
{\mathbf{x}^{k+1}_i=P_\Omega(\mathbf{x}^{k}_i-(1-\lambda)(\mathbf{x}^k_i-\text{arg}\min_{\mathbf{x}^k_j}\{\check{f}(\mathbf{x}^k_j)\})-\lambda
p\tau}),
\end{equation}
where $\lambda\in(0,1)$, $\tau$ is a vector with each element being
a random number, and  \begin{equation*}
p=\frac{\check{f}(\mathbf{x}^k_i+\tau)-\min\{\check{f}(\mathbf{x}^k_i+\tau)\}}{\max\{\check{f}(\mathbf{x}^k_i+\tau)\}-\min\{\check{f}(\mathbf{x}^k_i+\tau)\}}.
\end{equation*}
Besides, $P_\Omega$ is a projection function and make the updated
position satisfy the simple bound constraints, where
$\Omega=\{\mathbf{x}\in\mathbb{R}^d|l_i\leq x_i\leq
u_i,i=1,2\cdots,d\}$. The mathematical definition of
$P_\Omega(\mathbf{x})$ is
$P_\Omega(\mathbf{x})=\text{arg}~\min_{\mathbf{y}\in\Omega}\|\mathbf{y}-\mathbf{x}\|_2$
with $\|\cdot\|_2$ denoting the Euclidean norm.  The algorithm for
the evaluation of $P_\Omega(\mathbf{x})$ is given in Algorithm 2.

\begin{algorithm}[t]
\caption{Algorithm for the evaluation of $P_\Omega(\mathbf{x})$ with
$\mathbf{x}=[x_1,x_2,\cdots,x_d]^\text{T}$}
\begin{algorithmic}
\FOR{$i=1:d$} \IF{$x_1<l_i$} \STATE $x_i=l_i$ \ENDIF \IF{$x_i>u_i$}
\STATE $x_i=u_i$ \ENDIF \ENDFOR \RETURN
$\mathbf{y}=[x_1,x_2,\cdots,x_d]^\text{T}$
\end{algorithmic}
\end{algorithm}

\begin{algorithm}[t]
\caption{PSA for COPs}
\begin{algorithmic}
\STATE {\it Cost function} $\check{f}(\mathbf{x})$ as defined in
(\ref{eq.checkf}), ~$\mathbf{x}=[x_1,x_2,\cdots,x_d]^\text{T}$
\STATE {\it Generate initial position of porcellio scaber}
$\mathbf{x}^0_i~(i=1,2,\cdots,N)$ {\it according to
(\ref{eq.ini})}\STATE {\it Environment condition} $E_{\mathbf{x}}$
{\it at position} $\mathbf{x}$ {\it is determined by}
$\check{f}(\mathbf{x})$ \STATE {\it Set weighted parameter}
$\lambda$ {\it for decision based on aggregation and the propensity
to explore novel environments} \STATE {\it Set penalty parameter
$\gamma$ in $\check{f}(\mathbf{x})$ to a large enough value} \STATE
{\it Initialize $f_{*}$ to an extremely large value} \STATE {\it
Initialize each element of vector $\mathbf{x}_{*}\in\mathbb{R}^d$ to
an arbitrary value}
 \WHILE{$k<MaxStep$}
\STATE {\it Get the position with the best environment condition,
i.e.,}
$\mathbf{x}_b=\text{arg}\min_{\mathbf{x}^k_j}\{f(\mathbf{x}^k_j)\}$
{\it at the current time among the group of porcellio scaber}
\IF{$\min_{\mathbf{x}^k_j}\{f(\mathbf{x}^k_j)\}<f_*$} \STATE
$\mathbf{x}_*=\mathbf{x}_b$ \STATE
$f_*=\min_{\mathbf{x}^k_j}\{f(\mathbf{x}^k_j)\}$\ENDIF \STATE {\it
Randomly chose a direction
$\tau=[\tau_1,\tau_2,\cdots,\tau_d]^\text{T}$ to detect} \STATE {\it
Detect the best environment condition} $\min\{E_{\mathbf{x}}\}$ {\it
and worst environment condition} $\max\{E_{\mathbf{x}}\}$ {\it at
position} $\mathbf{x}^{k}_i+\tau$ {\it for} $i=1:N$ {\it all} $N$
{\it porcellio scaber} \FOR{$i=1:N$~{\it all $N$ porcellio scaber}}
\STATE {\it Determine the difference with respect to the position to
aggregate i.e.,}
$\mathbf{x}^k_i-\text{arg}\min_{\mathbf{x}^k_j}\{f(\mathbf{x}^k_j)\})$
 \STATE {\it Determine where to explore, i.e.}, $p\tau$
 \STATE {\it Move to a new position according to (\ref{eq.mod}) where} $P_\Omega(\mathbf{x})$ {\it is evaluated via Algorithm 2} \ENDFOR \ENDWHILE \STATE{\it Output $\mathbf{x}_*$ and
the corresponding function value $f_*$} \STATE {\it Visualization}
\end{algorithmic}
\end{algorithm}

\subsection{PSA for COPs}

Based on the above modifications, the resultant PSA for solving COPs
is given in Algorithm 3. In the following section, we will use some
benchmark problems to test the performance of the PSA in solving
COPs.

\section{Case Studies}\label{sec.4}

In this section, we present experiment results regarding using the
PSA for solving COPs.

\subsection{Case I: Pressure vessel  problem}

\begin{figure}[t]\centering
\includegraphics[scale=0.6]{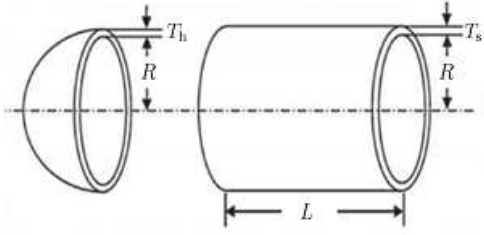}
\caption{A diagram showing the design parameters of a pressure
vessel \cite{fastf}. \label{fig.1}}
\end{figure}

\begin{algorithm}[t]
\caption{Algorithm for
$P_\Omega(\mathbf{x}=[x_1,x_2,x_3,x_4]^\text{T})$ in the pressure
vessel  problem}
\begin{algorithmic}
\STATE $y_1=round(x_1/0.0625)\times0.0625$ \IF{$y_1<0.0625$} \STATE
$y_1=0.0625$\ENDIF \IF{$y_1>99\times0.0625$} \STATE
$y_1=99\times0.0625$\ENDIF \STATE
$y_2=round(x_2/0.0625)\times0.0625$ \IF{$y_2<0.0625$} \STATE
$y_2=0.0625$\ENDIF \IF{$y_2>99\times0.0625$} \STATE
$y_2=99\times0.0625$\ENDIF \IF {$x_3<10$} \STATE $y_3=10$\ENDIF \IF
{$x_3>200$} \STATE $y_3=200$\ENDIF \IF {$x_4<10$} \STATE
$y_4=10$\ENDIF \IF {$x_4>200$} \STATE $y_4=200$\ENDIF \RETURN
$\mathbf{y}=[y_1,y_2,y_3,y_4]^\text{T}$
\end{algorithmic}
\end{algorithm}

\begin{table*}[t]
\centering
 \caption{\label{tab2}Comparisons of best results for the pressure
vessel problem}
 \begin{tabular}{llllllllllll}
  \toprule
  & $x_1$ & $x_2$ & $x_3$ & $x_4$ & $g_1(\mathbf{x})$ & $g_2(\mathbf{x})$ & $g_3(\mathbf{x})$ & $g_4(\mathbf{x})$ &$f(\mathbf{x})$\\
  \midrule
  \cite{csaam} & 0.8125 & 0.4375 &
  42.0984 & 176.6366 &8.00e-11$\dag$ & -0.0359 & -2.724e-4 &
  -63.3634 & 6059.7143\\
  \cite{fastf} & 0.7782 & 0.3846 & 40.3196 & 200.000 &
  -3.172e-5 & 4.8984e-5$\dag$ & 1.3312$\dag$ & -40 & 5885.33\\
  \cite{aipso} & 0.8125 & 0.4375 & 42.0984
  & 176.6366 & 8.00e-11$\dag$ & -0.0359 & -2.724e-4 &
  -63.3634 & 6059.7143\\
 \cite{anmha} &  1.125 & 0.625 & 58.2789 & 43.7549
&-0.0002 & -0.06902 & -3.71629 & -196.245 & 7198.433\\
\cite{niadp} & 1.125 & 0.625 & 48.97 & 106.72 & -0.1799 &
-0.1578 & 97.760 & -132.28 & 7980.894 \\
\cite{gafnm} & 1.125 & 0.625 & 58.1978 & 44.2930 &
-0.00178 &-0.06979 & -974.3 &-195.707 &7207.494\\
\cite{uoasa} & 0.8125 & 0.4375 & 40.3239 & 200.0000 &
-0.034324 & -0.05285 & -27.10585 & -40.0000 & 6288.7445\\
\cite{genas} & 0.9375 & 0.5000 & 48.3290 & 112.6790 & -0.0048
&-0.0389 &-3652.877&-127.3210&6410.3811\\
 \cite{aalmb} & 1.125 & 0.625 & 58.291 & 43.690 &
0.000016 & -0.0689 & -21.2201 & -196.3100 & 7198.0428\\
\cite{asbsm} & 0.8125 & 0.4375 & 41.9768 &
182.2845 & -0.0023 & -0.0370 & -22888.07 & -57.7155 & 6171.000\\
 \cite{gofsd}& 1.000 & 0.625 & 51.000 & 91.000 &
-0.0157 & -0.1385 & -3233.916 & -149 & 7079.037\\
 \cite{hagaw} & 0.8125 & 0.4375 & 42.0870 &
 176.7791 & -2.210e-4 &-0.03599 & -3.51084 & -63.2208 &
 6061.1229\\
 \cite{agafn} & 1 & 0.625 &51.2519 & 90.9913 & -1.011 &
-0.136 & -18759.75 &-149.009 & 7172.300\\

 \cite{aiaco} & 0.8125 & 0.4375 & 42.0984 &
176.6378 &-8.8000e-7 &-0.0359 &-3.5586 & -63.3622 &6059.7258\\

 PSA & 0.8125 & 0.4375 & 42.0952 &
176.8095 & -6.2625e-5 & -0.0359
&-738.7348 & -63.1905 &6063.2118\\
  \bottomrule
 \end{tabular}
 \begin{tablenotes}
\item[1] $\dag$ means that the corresponding constraint is violated.
\end{tablenotes}
\end{table*}

In this subsection, the pressure vessel  problem is considered. The
pressure vessel problem is to find a set of four design parameters,
which are demonstrated in Fig. \ref{fig.1}, to minimize the total
cost of a pressure vessel considering the cost of material, forming
and welding \cite{1}.  The four design parameters are the inner
radius $R$, and the length $L$ of the cylindrical section, the
thickness $T_\text{h}$ of the head, the thickness $T_\text{s}$ of
the body. Note that, $T_\text{s}$ and $T_\text{h}$ are integer
multiples of 0.0625 in., and $\text{R}$ and $L$ are continuous
variables.

Let
$\mathbf{x}=[x_1,x_2,x_3,x_4]^\text{T}=[T_\text{s},T_\text{h},R,L]^\text{T}$.
The pressure vessel  problem can be formulated as follows
\cite{fastf}:
\begin{equation*}
\begin{aligned}
\text{minimize}~f(\mathbf{x})&=0.6224x_1x_3x_4+1.7781x_2x^2_3\\
&~~+3.1661x^2_1x_4+19.84x^2_1x_3,\\
\text{subject to}~g_1(\mathbf{x})&=-x_1+0.0193x_3\leq0,\\
g_2(\mathbf{x})&=-x_2+0.00954x_3\leq0,\\
g_3(\mathbf{x})&=-\pi x^2_3x_4-\frac{4}{3}\pi x^3_3+1296000\leq0, \\
g_4(\mathbf{x})&=x_4-240\leq0, \\
x_1&\in\{1,2,3,\cdots,99\}\times0.0625,\\
x_2&\in\{1,2,3,\cdots,99\}\times0.0625,\\
x_3&\in[10,200],\\
x_4&\in[10,200].
\end{aligned}
\end{equation*}
Evidently, this problem has a nonlinear cost function, three linear
and one nonlinear inequality constraints. Besides, there are two
discrete and two continuous design variables. Thus, the problem is
relatively complicated. As this problem is a mixed
discrete-continuous optimization, the projection function
$P_\Omega(\mathbf{x})$ is slightly modified and presented in
Algorithm 4. Besides, the initialization of the initial positions of
porcellio scaber is modified as follows:
\begin{equation*}
\begin{aligned}
{x}^0_{i,1}&=0.0625+{floor}((99-1)\times
{rand})\times0.0625,\\
{x}^0_{i,2}&=0.0625+{floor}((99-1)\times
{rand})\times0.0625,\\
{x}^0_{i,3}&=10+{floor}(200-10)\times{rand},\\
{x}^0_{i,4}&=10+{floor}(200-10)\times{rand},
\end{aligned}
\end{equation*}
where ${x}^0_{i,j}$ denotes the $j$th variable of the position
vector of the $i$th porcellio scaber;
${floor}(y)=\text{arg}\text{min}_{x\in\{0,1,2,\cdots\}}\{x+1>y\}$,
i.e., the ${floor}$ function obtains the integer part of a real
number; ${rand}$ denotes a random number in the region $(0,1)$. The
functions ${floor}$ and ${rand}$ are available at Matlab.

The best result we obtained using the PSA in 1000 instances of
executions and those by using various existing algorithms or methods
for solving this problem are listed in Table \ref{tab2}. Note that,
in the experiments, 40 porcellio scaber are used, the parameter
$\lambda$ is set to 0.6, and the $MaxStep$ is set to 100000 with
$\tau$ being a zero-mean random number with the standard deviation
being 0.1. As seen from Table \ref{tab2}, the best result obtained
by using the PSA is better than most of the existing results.
Besides, the difference between the best function value among all
the ones in the table and the best function value obtained via using
the PSA is quite small.

\begin{table*}[t]
\centering
 \caption{\label{tab3}Comparisons of best results for Himmelblau's nonlinear optimization problem}
 \begin{tabular}{llllllllllll}
  \toprule
   & $x_1$ & $x_2$ & $x_3$ & $x_4$ & $x_5$ & $g_1(\mathbf{x})$ & $g_2(\mathbf{x})$ & $g_3(\mathbf{x})$ &$f(\mathbf{x})$\\
  \midrule

  \cite{couc} & 78.0 & 33.0 & 27.07997 & 45.0
  & 44.9692& 92.0000 & 100.4048 & 20.0000 & -31025.5602\\
  \cite{covga} & 78.00 & 33.00 & 29.995 & 45.00 &
  36.776 & 90.7147 & 98.8405 &19.9999$\dag$ & -30665.6088\\
  \cite{gaaed} & 81.4900 & 34.0900 & 31.2400 & 42.2000
  & 34.3700 & 90.5225 & 99.3188 & 20.0604 & -30183.576\\
  \cite{anlp} & 78.6200 & 33.4400 & 31.0700 & 44.1800 &
  35.2200 & 90.5208 & 98.8929 &20.1316 &-30373.949\\
  \cite{mveob} & 78.00 &33.00 &29.995256 &45.00
  &36.775813&92 &98.8405 &20&-30665.54\\
  PSA & 79.9377 & 33.8881 & 28.5029 & 41.3052 & 41.7704 & 91.6157 &
  100.4943 & 20.0055 & -30667.8113\\
  \bottomrule
 \end{tabular}
 \begin{tablenotes}
\item[1] $\dag$ means that the corresponding constraint is violated.
\end{tablenotes}
\end{table*}

\subsection{Case II: Himmelblau's nonlinear optimization problem}
In this subsection, we consider a nonlinear optimization problem
proposed by Himmelblau \cite{anlp}. This problem is also one of the
well known benchmark problems for bio-inspired algorithms. The
problem is formally described as follows \cite{anlp}:
\begin{equation*}
\begin{aligned}
\text{minimize}~f(\mathbf{x})=& 5.3578547x^2_3+0.8356891x_1x_5\\
&+37.29329x_1-40792.141,\\
\text{subject to}~g_1(\mathbf{x})=& 85.334407+0.0056858x_2x_5\\
&+0.00026x_1x_4-0.0022053x_3x_5,\\
g_2(\mathbf{x})=&80.51249+0.0071317x_2x_5\\
&+0.0029955x_1x_2+0.0021813x^2_3,\\
g_3(\mathbf{x})=&9.300961+0.0047026x_3x_5\\
&+0.0012547x_1x_3+0.0019085x_3x_4,\\
&0\leq g_1(\mathbf{x})\leq 92,\\
&90\leq g_2(\mathbf{x})\leq110,\\
&20\leq g_3(\mathbf{x})\leq25,\\
&78\leq x_1\leq102,\\
&33\leq x_2\leq45,\\
&27\leq x_3\leq45,\\
&27\leq x_4\leq45,\\
&27\leq x_5\leq45,\\
\end{aligned}
\end{equation*}
with $\mathbf{x}=[x_1,x_2,x_3,x_4,x_5]^\text{T}$ being the decision
vector. In this problem, each double-side nonlinear inequality can
be represented by two single-side nonlinear inequality constraints.
For example, the constraint $90\leq g_2(\mathbf{x})\leq110$ can be
replaced by the following two constraints:
\begin{equation*}
\begin{aligned}
-g_2(\mathbf{x})&\leq-90, \\
g_2(\mathbf{x})&\leq110.
\end{aligned}
\end{equation*}
Thus, this problem can also be solved by the PSA proposed in this
paper.

The best result we obtained via using the PSA in 1000 instances of
executions, together with the result obtained by other algorithms or
methods, is listed in Table \ref{tab3}. In the experiments, 40
porcellio scaber are used, the parameter $\lambda$ is set to 0.6,
and the $MaxStep$ is set to 100000 with $\tau$ being a zero-mean
random number with the standard deviation being 0.1. Evidently, the
best result generated by the PSA is ranked No. 2 among all the
results in Table \ref{tab3}.

By the above results, we conclude that the PSA is a relatively
promising algorithm for solving constrained optimization problems.
The quite smalle performance difference between the PSA and the best
one may be the result of the usage of the penalty method with a
constant penalty parameter.

\section{Conclusions}\label{sec.5}
In this paper, the bio-inspired algorithm PSA has been extended to
solve nonlinear constrained optimization problems by using the
penalty method. Case studies have validated the efficacy and
superiority of the resultant PSA. The results have indicated that
the PSA is a promising algorithm for solving constraint optimization
problems. There are several issues that requires further
investigation, e.g., how to select a best penalty parameter that not
only guarantees the compliance with constraints but also the
optimality of the obtained solution. Besides, how to enhance the
efficiency of the PSA is also worth investigating.

\section*{Acknowledgement}
This work is supported by the National Natural Science Foundation of
China (with numbers 91646114, 61370150, and 61401385), by Hong Kong
Research Grants Council Early Career Scheme (with number 25214015),
and also by Departmental General Research Fund of Hong Kong
Polytechnic University (with number G.61.37.UA7L).


\begin{thebibliography}{}




\bibitem{1} J. Kennedy and R. Eberhart, ``Particle swarm
optimization,'' in \emph{Proc. IEEE Int. Conf. Neural Netw.}, 1995,
1942--1948.

\bibitem{2}  A. H. Gandomi, X. Y. Yang, ``Benchmark problems in structural optimization,'' in \emph{Computational Optimization Methods And
Algorithms}, Berlin: Springer-Verlag, 2011, pp. 259--281.




\bibitem{aipso} S. He , E. Prempain, and Q. H. Wu, ``An improved particle swarm optimizer for
mechanical design optimization problems,'' \emph{Eng. Opt.}, vol.
36, no. 5, pp. 585--605, 2004.







\bibitem{csaam}
A. H. Gandomi, X. S. Yang, and A. H. Alavi, ``Cuckoo search
algorithm: A metaheuristic approach to solve structural optimization
problems,'' \emph{Eng. Comput.}, vol. 29, pp. 17--35, 2013.

\bibitem{doose} O. Pauline, H. C. Sin, D. D. C. V. Sheng, S. C. Kiong, and O. K.
M., ``Design optimization of structural engineering problems using
adaptive cuckoo search algorithm,'' in \emph{Proc. 3rd Int. Conf.
Control Autom. Robot.}, 2017, pp. 745--748.

\bibitem{eoclb} V. Garg and K. Deep, ``Effectiveness of constrained laplacian
biogeography based optimization for solving structural engineering
design problems,'' in \emph{Proc. 6th Int. Conf. Soft. Comput.
Problem Solv.}, 2017, pp. 206--219.


\bibitem{anlp}
D. Himmelblau, \emph{Applied Nonlinear Programming}, New York:
McGraw-Hill, 1972.

\bibitem{PSA}
Y. Zhang and S. Li, ``PSA: A novel optimization algorithm based on
survival rules of porcellio scaber,'' Available at
https://arxiv.org/abs/1709.09840.














\bibitem{fastf}
X. S. Yang, ``Firefly algorithm, stochastic test functions and
design optimisation,'' \emph{Int. J. Bio-Inspired Comp.}, vol. 2,
no. 2, pp. 2, pp. 78--84, 2010.

\bibitem{anmha}
K. S. Lee and Z. W. Geem, ``A new meta-heuristic algorithm for
continuous engineering optimization: Harmony search theory and
practice,'' \emph{Comput. Methods Appl. Mech. Engrg.}, vol. 194, pp.
3902--3933, 2005.

\bibitem{niadp}
E. Sandgren, ``Nonlinear integer and discrete programming in
mechanical design optimization,'' \emph{J. Mech. Des. ASME}, vol.
112, 223--229, 1990.

\bibitem{gafnm}
S. J. Wu and P. T. Chow, ``Genetic algorithms for nonlinear mixed
discrete-integer optimization problems via meta-genetic parameter
optimization,'' \emph{Engrg. Optim.}, vol. 24, 137--159, 1995.

\bibitem{uoasa}
C. A. C. Coello, ``Use of a self-adaptive penalty approach for
engineering optimization problems,'' \emph{Comput. Ind.}, vol. 41,
no. 2, pp. 113--127, 2000.



\bibitem{genas}
 K. Deb, ``GeneAS: a robust optimal design technique for
mechanical component design'', in: D. Dasgupta, Z. Michalewicz Eds.,
\emph{Evolutionary Algorithms in Engineering Applications},
Springer-Verlag, Berlin, 1997, pp. 497--514.

\bibitem{aalmb}
B. K. Kannan and S. N. Kramer, ``An augmented Lagrange multiplier
based method for mixed integer discrete continuous optimization and
its applications to mechanical design,'' \emph{J. Mecha. Design,
Trans. ASME}, vol. 116, pp. 318--320, 1994.

\bibitem{asbsm}
S. Akhtar, K. Tai, T. Ray, ``A socio-behavioural simulation model
for engineering design optimization,'' \emph{Eng. Optmiz.}, vol. 34,
no. 4, pp. 341--354, 2002.

\bibitem{gofsd}
J. F. Tsai, H. L. Li, and N. Z. Hu, ``Global optimization for
signomial discrete programming problems in engineering design,''
\emph{Eng. Optmiz.} no. 34, no. 6, pp. 613--622, 2002.


\bibitem{hagaw}
C. A. C. Coello and N. C. Corte\'s, ``Hybridizing a genetic
algorithm with an artificial immune system for global
optimization,'' \emph{Eng. Optmiz.}, vol. 36, no. 5, pp. 607--634,
2004.

\bibitem{agafn}
H. L. Li and C. T. Chou, ``A global approach for nonlinear mixed
discrete programming in design optimization,'' \emph{Eng. Optmiz.}
vol. 22, pp. 109--122 1994.


\bibitem{aiaco}
A. Kaveh and  S. Talatahari, ``An improved ant colony optimization
for constrained engineering design problems,'' \emph{Eng. Comput.}
vol. 27, no. 1, pp. 155--182, 2010.





\bibitem{couc}
G. H. M. Omran and A. Salman, ``Constrained optimization using
CODEQ,'' \emph{Chaos Soliton. Fract.}, vol. 42, pp. 662--668, 2009.

\bibitem{covga}
A. Homaifar, S. Lai, X. Qi, ``Constrained optimization via genetic
algorithms,'' \emph{Simulation}, vol. 62, no. 4, pp. 242--253, 1994.

\bibitem{gaaed}
M. Gen and R. Cheng, \emph{Genetic Algorithms and Engineering
Design}, Wiley, New York, 1997.

\bibitem{mveob}
G. G. Dimopoulos, ``Mixed-variable engineering optimization based on
evolutionary and social metaphors,'' \emph{Comput. Method. Appl.
Mech. Eng.}, vol. 196, pp. 803--817, 2007.




\end{thebibliography}
\end{document}